\documentclass[sigconf]{acmart}
\AtBeginDocument{%
  \providecommand\BibTeX{{%
    \normalfont B\kern-0.5em{\scshape i\kern-0.25em b}\kern-0.8em\TeX}}}

\copyrightyear{2021} 
\acmYear{2021} 
\setcopyright{acmlicensed}\acmConference[SSDBM 2021]{33rd International Conference on Scientific and Statistical Database Management}{July 6--7, 2021}{Tampa, FL, USA}
\acmBooktitle{33rd International Conference on Scientific and Statistical Database Management (SSDBM 2021), July 6--7, 2021, Tampa, FL, USA}
\acmPrice{15.00}
\acmDOI{10.1145/3468791.3468814}
\acmISBN{978-1-4503-8413-1/21/07}

\usepackage{graphicx}

\usepackage{amsmath}
\usepackage{nicefrac}

\usepackage{subcaption}
\usepackage{wrapfig}

\usepackage{algorithm, algpseudocode}

\usepackage{multirow}
\usepackage{array}

\usepackage{booktabs}
\usepackage{tabularx}
\usepackage{placeins}

\usepackage{color}

\DeclareMathOperator{\pool}{\text{pool}}

\DeclareMathOperator{\FF}{\mathcal{F}}
\DeclareMathOperator{\GG}{\mathcal{G}}
\DeclareMathOperator{\Loss}{\mathcal{L}}
\DeclareMathOperator{\RR}{\mathbb{R}}
\DeclareMathOperator{\Akipf}{\hat{\tilde{A}}}
\DeclareMathOperator{\Bkipf}{\hat{\tilde{B}}}

\newcommand{\softmax}[1]{\text{softmax} \left( #1 \right)}

\usepackage{balance} 

\begin{document}

\title{NF-GNN: Network Flow Graph Neural Networks for Malware Detection and Classification}

\author{Julian Busch}
\email{busch@dbs.ifi.lmu.de}
\affiliation{%
  \institution{LMU Munich}
  \city{Munich}
  \country{Germany}
}

\author{Anton Kocheturov}
\email{anton.kocheturov@siemens.com}
\affiliation{%
  \institution{Siemens Technology}
  \city{Princeton}
  \state{NJ}
  \country{USA}
}

\author{Volker Tresp}
\email{volker.tresp@siemens.com}
\affiliation{%
  \institution{Siemens AG}
  \city{Munich}
  \country{Germany}
}

\author{Thomas Seidl}
\email{seidl@dbs.ifi.lmu.de}
\affiliation{%
  \institution{LMU Munich}
  \city{Munich}
  \country{Germany}
}

\renewcommand{\shortauthors}{Busch, et al.}

\begin{abstract}
Malicious software (malware) poses an increasing threat to the security of communication systems as the number of interconnected mobile devices increases exponentially. 
While some existing malware detection and classification approaches successfully leverage network traffic data, they treat network flows between pairs of endpoints independently and thus fail to leverage rich communication patterns present in the complete network.
Our approach first extracts flow graphs and subsequently classifies them using a novel edge feature-based graph neural network model. We present three variants of our base model, which support malware detection and classification in supervised and unsupervised settings.
We evaluate our approach on flow graphs that we extract from a recently published dataset for mobile malware detection that addresses several issues with previously available datasets.
Experiments on four different prediction tasks
consistently demonstrate the advantages of our approach and show that our graph neural network model can boost detection performance by a significant margin. 
\end{abstract}



\maketitle

\section{Introduction}
Malicious software (malware) poses a significant threat to the security of information technology (IT) and operational technology (OT) in private and corporate environments. Along with an increasing degree of digitalization and the rise of new technologies such as the Internet of Things (IoT), an increasing number of devices, including mobile devices, such as smartphones, and Industrial Control Systems (ICS), become connected and thus are potential targets for attacks. Accurate detection and classification of malware is thus a vital task to ensure the security of such systems.

In this work, we focus on dynamic malware detection and classification. In contrast to static methods, which analyze a candidate application's source code or the structure of its executable, dynamic methods execute the application in a controlled environment and analyze dynamic behavior that can usually not be extrapolated from static data.
More specifically, we consider network traffic data generated during the execution of the application, which provides valuable insights into the dynamic and potentially malicious behavior affecting that application. However, in contrast to existing works, which classify individual network flows between two endpoints, i.e., aggregated communication between the two endpoints during some time frame, we construct a communication graph from all recorded network flows between any two endpoints during that time frame to obtain a rich representation of communication in the network. To the best of our knowledge, no existing work has considered this setting so far.

While classical machine learning methods have proven successful for malware detection and classification \citep{gibert2020rise}, deep learning approaches have not been studied as extensively. Deep learning methods offer an additional advantage of automatically learning suitable feature representations of the input data optimized for the task at hand, in contrast to traditional feature engineering approaches. Our novel edge feature-based graph neural network model learns suitable representations from extracted network flow graphs. Along with a base model for learning graph representations, we propose three derived model variants, a graph classifier, a graph autoencoder, and a one-class graph neural network, which are able to perform supervised malware detection and classification and unsupervised malware detection, respectively.

We evaluate our approach on a graph dataset that we extract from network flow features obtained from traffic generated by android applications executed on real mobile devices. The original flow features are provided by \citep{lashkari2018toward}. The data was collected in a carefully designed environment to account for several common defects observed in previously used datasets.
Instead of classifying individual network flows, as the original work proposes, we follow our graph-based approach and construct a flow graph 
for each execution of a candidate application. 
Experiments on four different prediction tasks, including supervised binary, category and family classification, and unsupervised detection, consistently demonstrate the significantly superior detection performance of our approach. Our neural network model additionally boosts performance compared to baseline models, even in settings with unlabeled data and small amounts of available training data.

We summarize our contributions as follows:

\begin{itemize}
\item We propose, to the best of our knowledge, the first graph-based approach to network traffic-based malware detection and classification.
\item We propose a novel method for extracting directed edge-attributed flow graphs from sets of network flows recorded in a monitored network.
\item We propose a novel graph neural network model that effectively learns representations from these graphs, utilizing the graph topology and edge attributes.
\item We provide an extensive experimental evaluation on a novel flow graph dataset considering four different supervised and unsupervised detection tasks.
\end{itemize}

\section{Related Work}
While static malware detection and classification methods analyze a candidate application's source code or the structure of its executable file, dynamic methods execute the application in a controlled environment and analyze its behavior. Such behavior is difficult and often impossible to extrapolate from static data. Further, static methods are commonly vulnerable to obfuscation techniques modifying the code or structure of the executable.
Our approach is a dynamic one, focusing on network traffic data generated by a particular candidate application during execution.

A comprehensive overview of existing machine learning methods for static and dynamic malware detection and classification is provided by a recent survey \cite{gibert2020rise}.
Methods relying on network traffic data mainly differ by which specific features are extracted and which machine learning algorithm is used. 
To the best of our knowledge, all existing approaches focus on classifying individual network traffic between two endpoints in a network, possibly at different resolutions. For instance, \citep{bekerman2015unknown} describes different resolution levels, ranging from single transactions over sessions to flows and conversations. Thereby, a flow describes traffic associated with a specific \texttt{(Source-IP, Source-Port, Destination-IP, Destination-Port, Protocol)}-tuple within a particular time frame and a conversation aggregates flows with the same Source- and Destination-IP endpoints.
Instead of focusing on bilateral flows or conversations between two endpoints, our approach extracts a graph from flows between any two endpoints in a monitored network during a specific time frame to obtain a richer representation of communication in the network. 

Existing methods 
typically report high performance on datasets that exhibit various common defects. The recently published \textit{CICAndMal2017} dataset \citep{lashkari2018toward} has addressed these issues by providing a sufficient number of malware samples from diverse malware categories and families with a realistic distribution of benign and malicious applications. Additionally, each application is executed on an actual physical device instead of an emulator or a virtual machine to accurately capture its actual behavior. 
Instead of classifying network flows individually as proposed by the authors, we instead extract network flow graphs from the given network traffic features. 

The results reported in \citep{lashkari2018toward} have been improved by adding new dynamic API-call features and combining a dynamic prediction model with a static prediction model \citep{taheri2019extensible}. Our model could be combined with a static prediction model in a similar fashion.
Improved accuracy could also be achieved by predicting labels for conversations instead of individual flows \citep{abuthawabeh2020enhanced}, further mitigating the effect of port randomization techniques. Our approach takes an additional step ahead by classifying flow graphs modeling the communication between all endpoints instead of only pairs of endpoints.
Two further works \cite{bibi2019effective,chen2019android} reported malware detection results only for a subset of the whole dataset.

Existing graph-based approaches to malware detection and classification mostly focus on static analysis, considering function call graphs \citep{kinable2011malware,hassen2017scalable,jiang2018dlgraph} or control graphs \citep{faruki2012mining,bazrafshan2013survey}.
Dynamic graph-based approaches include \citep{anderson2011graph}, where graphs are constructed from instruction traces collected during execution, and \citep{john2020graph}, where system call graphs are considered. 
The latter work is most related to our approach since a \emph{Graph Convolutional Neural Network (GCN)} \citep{kipf2016semi} is used to learn node features. 
To the best of our knowledge, no existing work has considered graph-based approaches based on dynamically generated network traffic data.

Deep learning methods have not been investigated as extensively as classical machine learning methods for network traffic-based approaches. 
Existing methods consider detection of endpoints generating malicious traffic \citep{prasse2017malware}, intrusion detection \citep{muna2018identification}, or ransomware detection \cite{bibi2019effective}.
To the best of our knowledge, no existing deep learning-based malware detection and classification method considers network flow graphs.

While our proposed approach includes a novel graph neural network model for malware detection and classification, it could be more generally employed to solve other graph classification and graph anomaly detection tasks. 
Graph neural networks \citep{bronstein2017geometric,gilmer2017neural,battaglia2018relational,wu2019comprehensive,kipf2016semi,velivckovic2017graph,xu2018powerful,wu2019simplifying,busch2020pushnet}
have recently become a de-facto standard for machine learning on graphs.
The majority of these methods can be described in a message-passing framework, as detailed in \citep{gilmer2017neural}. The main idea is to iteratively propagate feature vectors through the graph by passing messages between nodes such that a node's representation depends on feature vectors appearing in its neighborhood.
While most existing graph neural network models consider only node attributes, our model focuses on edge attributes instead. 
Some existing models consider edge attributes \citep{simonovsky2017dynamic,kipf2018neural,schlichtkrull2018modeling,gong2019exploiting,jiang2020co,wu2020comprehensive}, but none of these models is directly applicable to our setting since they either consider multi-relational graphs or focus on node attributes and only utilize edge features to improve message passing between nodes. Other existing models \citep{shang2018edge,zhang2018graph,lin2020graph} focus on more specific tasks which differ from our setting.

\begin{figure*}[t!]
    \centering
    \begin{subfigure}[t]{0.64\textwidth}
		\centering
		\includegraphics[width=\textwidth]{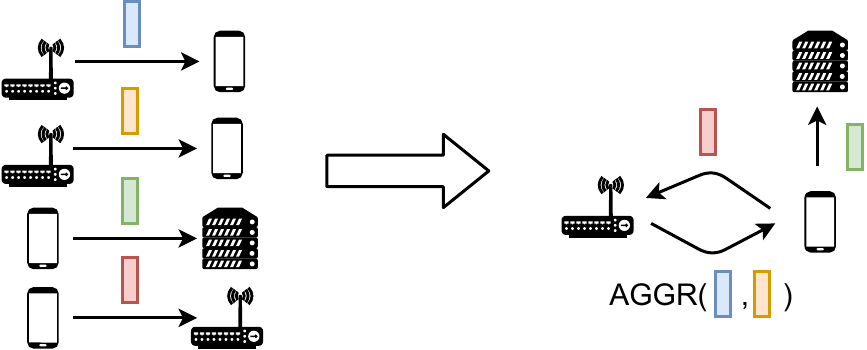}
		\caption{Network Flow Graph Extraction}\label{fig:graph_extraction}
	\end{subfigure}
	\begin{subfigure}[t]{0.32\textwidth}
		\centering
		\includegraphics[width=\textwidth]{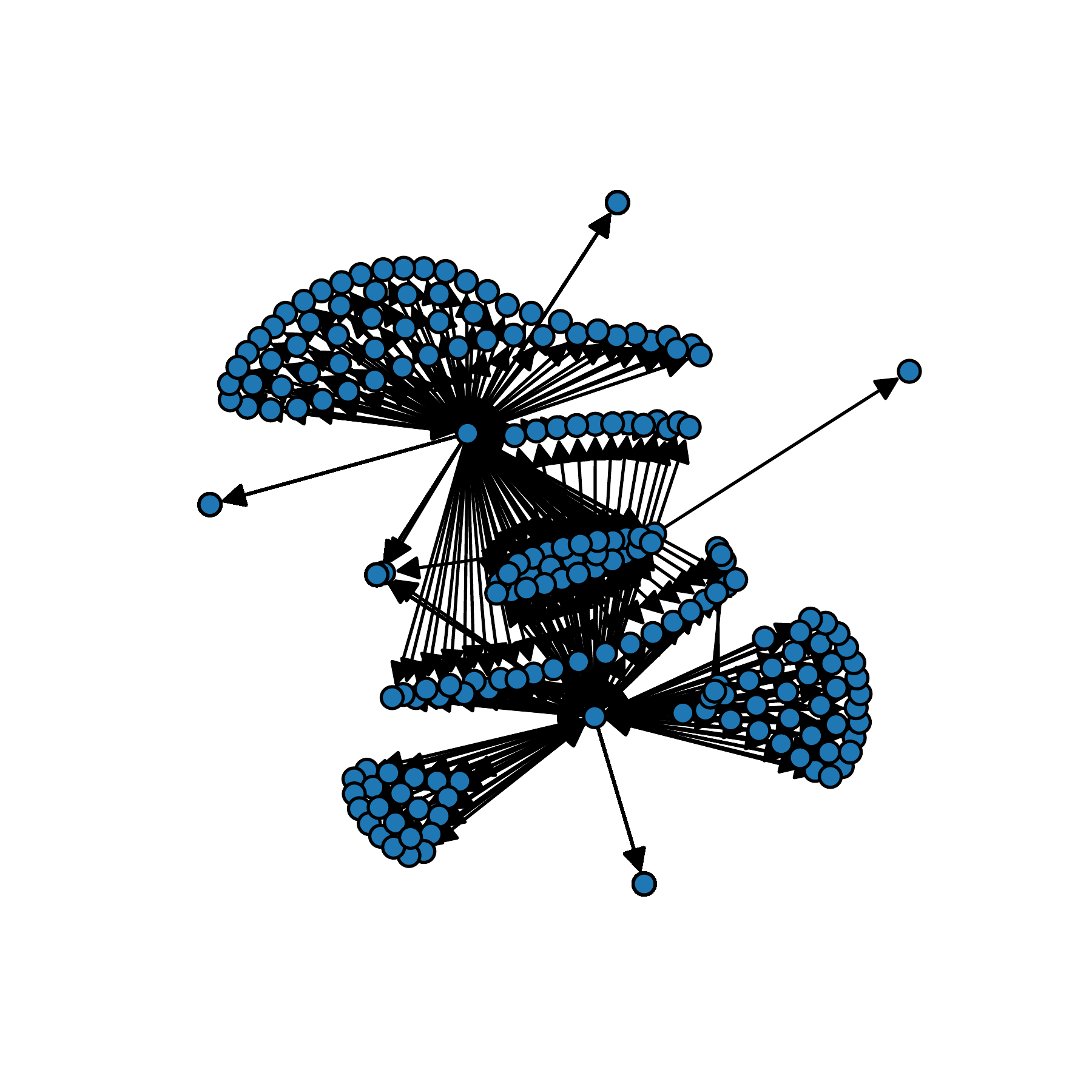}
		\caption{Real Network Flow Graph}\label{fig:graph_scareware}
	\end{subfigure}
    \caption{From a set of network flows recorded during a specific time frame, a flow graph is constructed by adding directed edges between all pairs of endpoints that communicated with at least one flow. Each flow is described by a feature vector summarizing its corresponding network traffic. Edges in the flow graph are annotated with these feature vectors, where feature vectors of parallel flows are aggregated. The topology of a real flow graph extracted from network traffic generated during execution of a \textit{FakeAV} Scareware application is shown in (\ref{fig:graph_scareware}).}
    \label{fig:graph}
\end{figure*}

\section{Network Flow Graph Extraction}
\label{sec:graph_extraction}

To decide whether a particular candidate instance of an application is malicious, we collect all network traffic generated during the execution of that application instance within a given time interval after installation.
The resulting data consists of a set of network flows described by feature vectors that can be extracted from \texttt{pcap}-files using tools such as \emph{CICFlowMeter} \citep{draper2016characterization}. Thereby, each flow $F$ describes network traffic associated with one particular \texttt{(Source-IP, Source-Port, Destination-IP, Destination-Port, Protocol)}-tuple during the considered time frame and has a feature vector $f \in \RR^d$ attached to it. Typical features include the number of packets sent, mean and standard deviation of the packet length, or minimum and maximum interarrival time of the packets.

From the resulting set of flows $\FF$, we extract a flow graph, where the nodes correspond to endpoints in the network and edges model communication between these endpoints. Instead of considering \texttt{(IP, Port)}-tuples, we factor out the port information and consider IP-endpoints for two main reasons:

\begin{enumerate}
    \item Apart from standard ports, port selection is often arbitrary and could even be subject to port randomization techniques, leading to arbitrary and potentially misleading graph structure.
    \item Empirically, we found \texttt{(IP, Port)}-graphs to be very sparse and rather uninformative. In comparison, IP-graphs exhibit much more interesting topologies.
\end{enumerate}

More specifically, from a set of flows $\FF$, we extract a directed graph $G=(V,E)$ where the nodes correspond to endpoints involved in any flow $F \in \FF$ and a directed edge is added for all pairs $(s_i, t_i)$ for which there exists a flow $F_i \in \FF$ with source and target IP $s_i$ and $t_i$, respectively. The feature vector assigned to this edge aggregates the feature vectors $f_i \in \RR^d$ of all flows $F_i$ along this edge using a set of five aggregation functions.
For each feature, the shape of the distribution of values along the edge is described using the first four moments, namely the mean, standard deviation, skew and kurtosis, and the median value. The aggregate values are computed for each feature and then concatenated, resulting in a feature vector $x_i \in \RR^{5d}$ for each edge $e_i \in E$. 
The flow graph extraction procedure is illustrated in Figure \ref{fig:graph_extraction}. Figure \ref{fig:graph_scareware} shows an exemplary graph extracted from real data.

Intuitively, the resulting graph captures how network traffic flows between different endpoints in the monitored network during a specific time frame. The graph structure reveals where traffic is flowing, and the additional edge attributes describe how it is flowing. Connecting individual flows in a graph provides a much richer relational representation compared to treating flows individually. Thus, we expect models learning from these graphs to perform significantly better at detection and classification tasks than models which classify individual flows. Our experimental results confirm this intuition. 

We wish to note that such graphs could potentially be used for further applications, such as intrusion detection or identifying devices generating malicious traffic.

\section{Network Flow Graph Neural Networks}

\begin{figure*}[t!]
	\centering
	\begin{subfigure}[t]{0.29\textwidth}
		\centering
		\includegraphics[width=.95\textwidth]{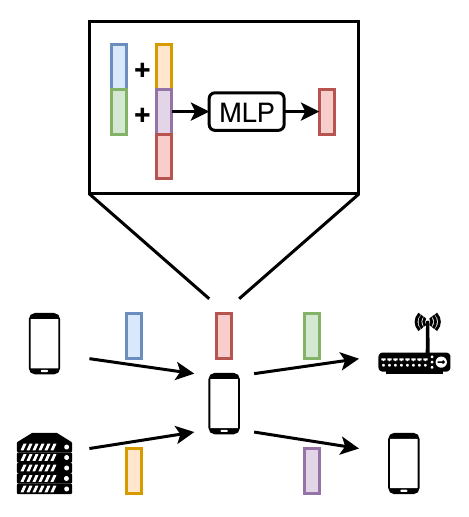}
		\caption{Node Update}\label{fig:node_update}
	\end{subfigure}
	\begin{subfigure}[t]{0.29\textwidth}
		\centering
		\includegraphics[width=\textwidth]{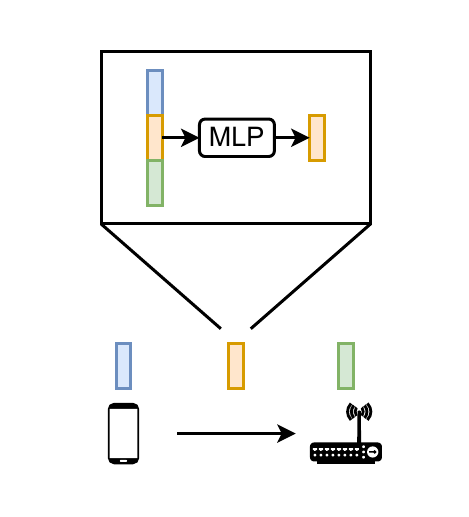}
		\caption{Edge Update}\label{fig:edge_update}
	\end{subfigure}
	\begin{subfigure}[t]{0.39\textwidth}
		\centering
		\includegraphics[width=\textwidth]{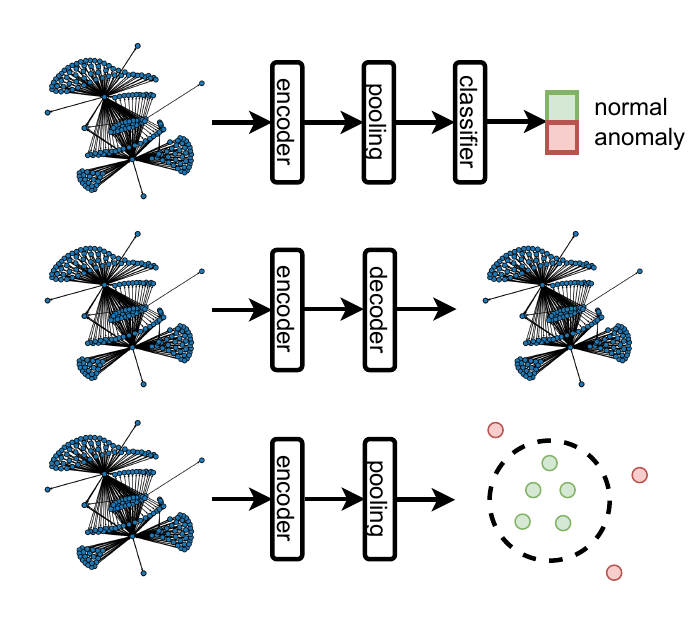}
		\caption{Model Variants}\label{fig:architectures}
	\end{subfigure}
	\caption{
	Our model learns how endpoints in the network communicate with each other by sequentially updating node (\ref{fig:node_update}) and edge feature vectors (\ref{fig:edge_update}) by neural message passing. The learned representations are used by our three model variants, a graph classifier, a graph autoencoder, and a graph one-class neural network, to solve supervised and unsupervised detection tasks (\ref{fig:architectures}).}
	\label{fig:graphs}
\end{figure*}

From a machine learning perspective, we consider two different problems: Supervised graph classification and unsupervised graph anomaly detection. To solve these problems, we introduce a novel graph neural network model, which learns expressive representations from network flow graphs, along with three different variants of that base model, a supervised graph classifier, an unsupervised graph autoencoder, and an unsupervised one-class graph neural network. The different model variants are illustrated in Figure \ref{fig:architectures}.

While we focus on malware detection and classification in this paper, the proposed models can be employed for general learning problems on (directed) graphs with edge attributes.

\subsection{Problem Setting}

Formally, we are given a set of graphs $\GG = \{G_1, \dots, G_N\}$, a set of edge feature matrices $X_i \in \RR^{m_i \times d}$, $i = 1, \dots, N$, where $m_i = |E_i|$, and a label matrix $\hat{Y} \in \{0, 1\}^{N \times c}$ providing a class label for each graph, where $c$ is the number of classes. 
For the sake of simplicity, we restrict ourselves to directed graphs in the following, though our models can be applied to undirected graphs in a straightforward fashion.
Further, it is not required that the input graphs share the same topology, 

In a supervised anomaly detection setting, class labels could be binary (normal vs. anomalous) or multi-class (normal class and different categories or families of anomalies). 
For supervised classification, a labeled training dataset is given as described above, and the task is to train a model which accurately predicts labels for new graphs not seen by the model during training.
For unsupervised anomaly detection, the training set is not labeled and consists of normal data and (usually a relatively small fraction of) anomalous data. The model is required to learn a concept of normality from the training data and correctly classify new graphs as either being normal or anomalous.

\subsection{Learning Representations of Network Flow Graphs}

Each input instance for our model is a directed graph $G=(V,E)$ with adjacency matrix $A \in \{0, 1\}^{n \times n}$ and an edge feature matrix $X \in \RR^{m \times d}$, where $m := |E|$. The representation learning part of our model computes latent representations of the edges and nodes in the graph and finally outputs a latent feature vector $h^{(1)} \in \RR^{h}$ for each node in the graph. Such a vector intuitively describes how the corresponding endpoint interacts with other endpoints in the network. Depending on the availability of labels, variants of our model compute either predictions or an anomaly score for an input graph from its latent node feature vectors. The model is trained end-to-end such that the latent representations are optimized towards the specific task. 
Given an input graph with edge attributes, our model performs the following feature transformation and propagation steps to sequentially compute latent representations of the graph's nodes and edges:

\begin{align}
\label{eq:e0}
E^{(0)} &= f_1(X) & & \in \RR^{m \times h} \\
\label{eq:h0}
H^{(0)} &= f_2 \left( \left[ \Bkipf_{in} E^{(0)}, \Bkipf_{out} E^{(0)} \right] \right)  & & \in \RR^{n \times h} \\
\label{eq:e1}
E^{(1)} &= f_3 \left( \left[ \Bkipf_{in}^T H^{(0)}, \Bkipf_{out}^T H^{(0)}, E^{(0)} \right] \right) & & \in \RR^{m \times h} \\
\label{eq:h1}
H^{(1)} &= f_4 \left( \left[ \Bkipf_{in} E^{(1)}, \Bkipf_{out} E^{(1)}, H^{(0)} \right] \right) & & \in \RR^{n \times h},
\end{align}

where $[\cdot, \cdot]$ denotes concatenation and $f_1, \dots, f_4$ are Multi-Layer Perceptrons (MLPs) with appropriate input and output dimensions. As per default, we use single-layer MLPs 
\begin{equation}
    f_i(X) = q\left( X W_i + b_i \right),
\end{equation}
where $W_i$ and $b_i$ are the learnable parameters of the model and $q$ is a non-linear activation. We use \emph{ReLU} activations and add batch normalization. The propagation matrices $\Bkipf_{in}, \Bkipf_{out} \in \RR^{n \times m}$ are obtained from the node-edge incidence matrices $B_{in}, B_{out} \in \{0, 1\}^{n \times m}$ with
\begin{equation}
(B_{in})_{ij} = \begin{cases}
1 & \text{if} \quad \exists v_k \in V: e_j = (v_k, v_i) \\
0 & \text{else} 
\end{cases}
\end{equation}
and	
\begin{equation}
(B_{out})_{ij} = \begin{cases}
1 & \text{if} \quad \exists v_k \in V: e_j = (v_i, v_k) \\
0 & \text{else} 
\end{cases},
\end{equation}
indicating in- and out-going edges for each node, by substituting non-zero entries with normalized edge weights.
Normalization is applied to preserve the scale of the feature vectors. In particular, we apply symmetric normalization to the adjacency as in \cite{kipf2016semi} before computing the node-edge incidence matrices, where the normalized adjacency matrix is given as $\Akipf = \tilde{D}^{-\nicefrac{1}{2}} \tilde{A} \tilde{D}^{-\nicefrac{1}{2}}$ with $\tilde{A} = A + I$ and degree matrix $\tilde{D}$.
Self-loops added for normalization are removed again such that the graph structure remains unchanged.
Illustrations of the performed node and edge feature updates are provided in Figure \ref{fig:node_update} and \ref{fig:edge_update}, respectively.

The first network layer learns how endpoints interact with each other directly by first applying a learnable feature transformation to the original edge feature vectors (Equation \ref{eq:e0}) and subsequently computing node representations by aggregating feature vectors from neighboring edges (Equation \ref{eq:h0}).
Notably, incoming and outgoing traffic is modeled separately for each node.

The second layer enables the model to learn how endpoints communicate indirectly with their 2-hop neighbors. In a first step, the edge features are updated again by transforming the concatenated feature vectors of the source and destination node and the edge features from the previous layer (Equation \ref{eq:e1}). Concatenating the edge features from the previous layer as residual connections \citep{he2016deep} gives this layer direct access to previously learned features and aids in optimization. Such skip-connections have shown to improve the performance of graph neural networks when applied to node features \citep{xu2018representation}, motivating us to apply them to edge features as well.
The edge feature update is followed by an update of the node features using features of incoming and outgoing edges and skip-connected node features from the first layer (Equation \ref{eq:h1}). These node representations constitute the final output of our representation learning module. 

In principle, one could add more layers to the model in a similar fashion to model interaction between more distant endpoints. 
However, the flow graphs considered in this paper usually have a relatively small diameter, such that additional layers might not result in improved performance but rather lead to over-fitting. In our experiments, we observed the best performance with either one or two layers.

\subsection{Network Flow Graph Classifier}
\label{sec:supervised}

For supervised graph classification, we append two more layers to the representation learning module. First, a pooling layer aggregates all node feature vectors in the input graph to a single vector describing the whole graph. The second layer predicts the graph label from the pooled graph representation,
\begin{align}
h &= \pool \left( H^{(1)} \right) & & \in \RR^h \label{eq:pool}\\
y &= \softmax{Wh + b} & & \in \RR^c  . 
\end{align}
Above, $\pool$ denotes a pooling function, such as element-wise mean or maximum, which aggregates all node representations into a single embedding vector for the whole graph. Predictions are computed by a dense prediction layer with learnable parameters $W \in \RR^{c \times h}$ and $b \in \RR^c$ and a softmax activation. 
The model parameters are then optimized w.r.t. the \emph{cross-entropy} loss
\begin{equation}
    \Loss_{CLF}(\GG) = \frac{1}{N \cdot c} \sum_{\substack{G_i \in \GG \\ j \in 1,\dots,c}} 
    - y_{ij} \log\hat{y}_{ij}.
\end{equation}
We denote this model as \emph{NF-GNN-CLF}.

\subsection{Network Flow Graph Autoencoder}
\label{sec:semi-supervised}

For unsupervised anomaly detection, autoencoder models often perform well in practice \citep{chalapathy2019deep}. In general, an autoencoder consists of two neural network modules.
While an encoder learns compact and expressive representations of the model inputs, a decoder is supposed to reconstruct the original inputs from their learned representations. If the model is trained with exclusively or mostly normal data, the reconstruction loss can be interpreted as an anomaly score, where instances incurring a larger reconstruction loss are considered more anomalous.

We propose a graph autoencoder model where our representation learning module acts as an encoder. The latent node representations $H^{(1)}$ are then used to reconstruct the original edge feature vectors of the graph using a decoder, which is a mirrored version of the encoder:

\begin{align}
\label{eq:dec_e1}
E^{(2)} &= f_5 \left( \left[ \Bkipf_{in}^T H^{(1)}, \Bkipf_{out}^T H^{(1)} \right] \right) & & \in \RR^{m \times h} \\
\label{eq:dec_h1}
H^{(2)} &= f_6 \left( \left[ \Bkipf_{in} E^{(2)}, \Bkipf_{out} E^{(2)}, H^{(1)} \right] \right)  & & \in \RR^{n \times h} \\
E^{(3)} &= f_7 \left( \left[ \Bkipf_{in}^T H^{(2)}, \Bkipf_{out}^T H^{(2)}, E^{(2)} \right] \right) & & \in \RR^{m \times h} \\
E^{(4)} &= f_8\left(E^{(3)}\right) & & \in \RR^{m \times h}
\end{align}

If the encoder uses only a single layer, the first layer of the decoder (Equation \ref{eq:dec_e1}--\ref{eq:dec_h1}) is dropped and the node embeddings $H^{(0)}$ are used as input instead. The model parameters are optimized w.r.t. a reconstruction loss
\begin{equation}
\Loss_{AE}(\GG) = \frac{1}{N} \sum_{G_i \in \GG} \frac{1}{m_i} \left\lVert X_i - E^{(4)}_i \right\rVert_F^2,
\end{equation}
where $||\cdot||_F$ denotes the Frobenius norm. A similar loss was used in \citep{cen2020anae} to reconstruct node attributes, whereas our model operates on edge-attributed graphs. We denote this variant of our model as \emph{NF-GNN-AE}.

\subsection{One-Class Network Flow Graph Neural Network}

Though autoencoder models perform well in practice, they don't optimize an anomaly detection objective directly. \emph{Deep SVDD} \citep{ruff2018deep} combines \emph{Support Vector Data Description (SVDD)} \citep{tax2004support} with a neural network for anomaly detection in a learned latent space. The main idea is to learn a transformation into a latent space such that most instances are mapped into a hypersphere in that space, and anomalous instances will fall outside of the hypersphere.

We propose a one-class graph neural network consisting of our representation learning module and an additional pooling layer, which summarizes each input graph into a single feature vector, similarly as for the supervised graph classifier.
The model parameters are optimized w.r.t. a one-class loss
\begin{equation}
    \Loss_{OC}(\GG) = \frac{1}{N} \sum_{G_i \in \GG} \left\lVert h_i - \mu \right\rVert_2^2 + \frac{\lambda}{2} \sum_{i=1}^4 \left\lVert W_i \right\rVert_F^2,
\end{equation}
where $\mu \in \RR^h$ denotes the center of the hyper-sphere in latent space. The center is initialized with the mean embedding vector of all graphs in $\GG$ after the first forward-pass and does not change thereafter. The second part of the loss regularizes the model parameters to limit model complexity. Bias vectors have been removed from all layers to prevent trivial solutions \citep{ruff2018deep}. We denote this variant of our model as \emph{NF-GNN-OC}.

\section{Experiments}

\begin{table*}[t!]
	\centering
	\resizebox{\textwidth}{!}{%

		\begin{tabular}{p{0.24\textwidth}llllll}
\toprule
{} & \multicolumn{3}{c}{\textbf{Weighted Recall}} & \multicolumn{3}{c}{\textbf{Weighted Precision}} \\
{} &            \textbf{Binary} &          \textbf{Category} &             \textbf{Family} &            \textbf{Binary} &          \textbf{Category} &             \textbf{Family} \\
\midrule
\textbf{Flows \cite{lashkari2018toward}} &  $88.30$ &  $48.50$ &  	$25.50$ &  $85.80$ &  $49.90$ &  $27.50$ \\
\textbf{Flows + Static + API Calls \cite{taheri2019extensible}} &  $95.30$ &  $81.00$ &  $61.20$ &  $95.30$ &  $83.30$ &  $59.70$ \\
\textbf{Conversations \cite{abuthawabeh2020enhanced}} &  $89.00$ &  $79.64$ &  $66.59$ &  $86.65$ &  $80.20$ &  $67.21$ \\
\midrule
\textbf{\textbf{SVM-LIN}   } &  $96.26 \pm 2.12$ &  $72.83 \pm 10.03$ &  $28.74 \pm 15.56$ &  $96.72 \pm 1.49$ &  $87.31 \pm 0.80$ &   $90.35 \pm 0.37$ \\
\textbf{\textbf{SVM-RBF}   } &  $96.20 \pm 1.51$ &   $85.42 \pm 3.84$ &  $49.96 \pm 16.20$ &  $96.64 \pm 1.06$ &  $89.13 \pm 2.18$ &   $91.52 \pm 0.98$ \\
\textbf{\textbf{KNN}       } &  $95.71 \pm 2.05$ &  $78.10 \pm 12.29$ &  $42.53 \pm 17.86$ &  $95.75 \pm 1.96$ &  $87.08 \pm 2.40$ &   $90.92 \pm 0.76$ \\
\textbf{\textbf{DT}        } &  $92.65 \pm 4.13$ &  $73.42 \pm 12.19$ &  $45.98 \pm 30.81$ &  $93.79 \pm 2.80$ &  $85.63 \pm 6.59$ &  $85.72 \pm 18.57$ \\
\textbf{\textbf{RF}        } &  $95.85 \pm 2.00$ &   $84.30 \pm 8.24$ &  $56.06 \pm 19.60$ &  $96.24 \pm 1.60$ &  $90.32 \pm 2.60$ &   $91.67 \pm 0.99$ \\
\textbf{\textbf{ADA}       } &  $96.38 \pm 1.62$ &  $76.02 \pm 12.67$ &  $42.17 \pm 28.82$ &  $96.67 \pm 1.31$ &  $83.97 \pm 6.72$ &  $87.88 \pm 13.35$ \\
\textbf{\textbf{MLP}       } &  $97.19 \pm 1.19$ &   $85.69 \pm 4.46$ &  $49.56 \pm 17.75$ &  $97.29 \pm 1.07$ &  $89.28 \pm 2.49$ &   $90.23 \pm 4.32$ \\
\midrule
\textbf{\textbf{NF-GNN-CLF}} &  $\mathbf{99.42 \pm 0.45}$ &   $\mathbf{95.41 \pm 1.48}$ &   $\mathbf{91.37 \pm 8.39}$ &  $\mathbf{99.44 \pm 0.44}$ &  $\mathbf{96.14 \pm 1.07}$ &   $\mathbf{93.62 \pm 2.52}$ \\
\bottomrule
\end{tabular}

	}
	\vspace{1em}
	\caption{Weighted recall and precision scores for the three supervised tasks. For competing methods, we report the scores provided by the respective authors. For our baselines and our model, scores are reported in terms of mean and standard deviation over 30 independent data splits.}
	\label{tab:results_supervised}
\end{table*}

We evaluate our approach on a graph dataset that we extract from the network traffic data provided by the \textit{CICAndMal2017} dataset \citep{lashkari2018toward}.
We focus on this datasets since it addresses several common defects shared by other existing datasets, allowing for a realistic and meaningful evaluation. These issues are addressed by providing a sufficient number of malware samples from diverse malware categories and families with a realistic distribution of benign and malicious applications. To accurately capture dynamic behavior, each application is executed on an actual physical device instead of an emulator or a virtual machine.

\subsection{Dataset Preparation}

Our extracted dataset consists of 2126 samples, where each sample corresponds to one instance of an android application installed and executed on a mobile phone. For each sample, all network flows within the network during execution are captured. For each flow, 80 features are recorded, including, e.g., the number of packets sent, mean and standard deviation of the packet length, and minimum and maximum interarrival time of the packets.
For a more detailed description of the data collection process, we refer to \citep{lashkari2018toward}. 

For each sample, 3 different labels are available, a binary label indicating whether the application is malicious or not, a category label with 5 possible values indicating the general class of malware, and a family label with 36 different values indicating the specific type of malware. Malware families with fewer than 9 samples have been removed to ensure a reasonable split into train, validation, and test sets. Consequently, for the family prediction task, only 2071 samples are available.

For each sample, we extract a graph as described in Section \ref{sec:graph_extraction} and remove the flow-id, timestamp, and endpoint IP and port information from the feature set. Additionally, we remove all features that are constant among all edges of all graphs, leaving 330 edge features.

To be able to compare against baselines apart from flow- and conversation-based methods and our graph neural network models, we construct additional datasets, which represent each sample by a single feature vector instead of a graph. We consider three different feature sets: 

\begin{enumerate}
    \item \textbf{Flow Features} For each sample, we aggregate the features of all flows for this sample using the same aggregation functions we used for deriving the edge feature vectors and concatenate the aggregates to a 318-dimensional feature vector.
    \item \textbf{Graph Features} To evaluate the importance of the graph topology for the baseline models, we extract a set of structural features from each graph. In particular, we extract 2 global features (global clustering coefficient and assortativity) and 8 local node-features (degree, number of 2-hop neighbors, clustering coefficient, avg. neighbor degree, avg. neighbor clustering coefficient, number of edges in egonet, number of edges leaving egonet, betweenness centrality) that are aggregated over all nodes in the graph, again using the same aggregation functions. All extracted graph features are concatenated, leading to a 40-dimensional feature vector.
    \item \textbf{Combined Features} To provide access to both types of features, we concatenate the flow and graph features to a combined 358-dimensional feature vector for each sample.
\end{enumerate}

Again, all constant feature columns have been removed. All feature matrices, including the edge feature matrices for our model, are standardized before training.

\subsection{Supervised Malware Detection and Classification}

We consider three supervised tasks, binary, category, and family classification. To ensure a fair and unbiased comparison, we follow a rigorous evaluation protocol.

\subsubsection{Experimental Setup}

First, we split the dataset into a train, validation, and test part. Each model is trained on the training set, hyper-parameters are chosen based on validation set performance using a grid search, and results are reported on the test set using the optimal hyper-parameter values. All experiments are repeated on 30 randomly generated splits, and mean and standard deviation of the results are reported. 

To ensure a balanced training set, we sample 100, 25, and 5 samples per class for training for binary, category, and family classification, respectively. For binary and category classification, 5\% of the remaining samples are sampled in a stratified fashion for validation. The remaining samples are used for testing. For family prediction, 20\% of the non-training samples are chosen for validation to account for smaller class sizes. Stratification ensures that smaller classes are represented appropriately in the validation set. 

Again, to account for class imbalance, we report weighted precision, recall, and F1 scores. In all considered settings, we checked that our models outperform the respective best competitor with statistical significance at $P < 1e-5$ according to a Wilcoxon signed-rank test.

We compare our model against 7 baseline algorithms with different inductive bias, \emph{Support Vector Machine (SVM)} with linear and RBF-kernel, \emph{k-Nearest Neighbor Classifier (KNN)}, \emph{Decision Tree (DT)}, \emph{Random Forest (RF)}, \emph{Adaboost (ADA)} and a \emph{Multi-Layer Perceptron (MLP)} with up to two dense layers and \emph{ReLU} and \emph{softmax} activations.
Additionally, we compare our graph-based approach against two existing flow-based approaches \cite{lashkari2018toward,taheri2019extensible} and one conversation-based approach \cite{abuthawabeh2020enhanced}. Since no source code has been made available, we report the scores provided by the respective authors. Results have been provided only for a single data split.

For all neural network models, we use early stopping based on validation set performance using a patience of 20 epochs and a maximum of 1000 epochs. The default learning rate for all MLP-models is fixed as $1e-3$. For our model, we additionally apply dropout regularization before each dense layer and on the propagation matrices. All considered hyper-parameter values for all models are provided in Appendix \ref{app:hyper_parameters}.

Since the original feature vectors are rather high-dimensional, we introduce an additional hyper-parameter for each baseline model indicating whether or not to perform \emph{Principle Component Analysis (PCA)} as feature reduction before training where 95\% of the variance in the data is retained.

\begin{figure*}[t]
	\centering
	\includegraphics[width=\textwidth]{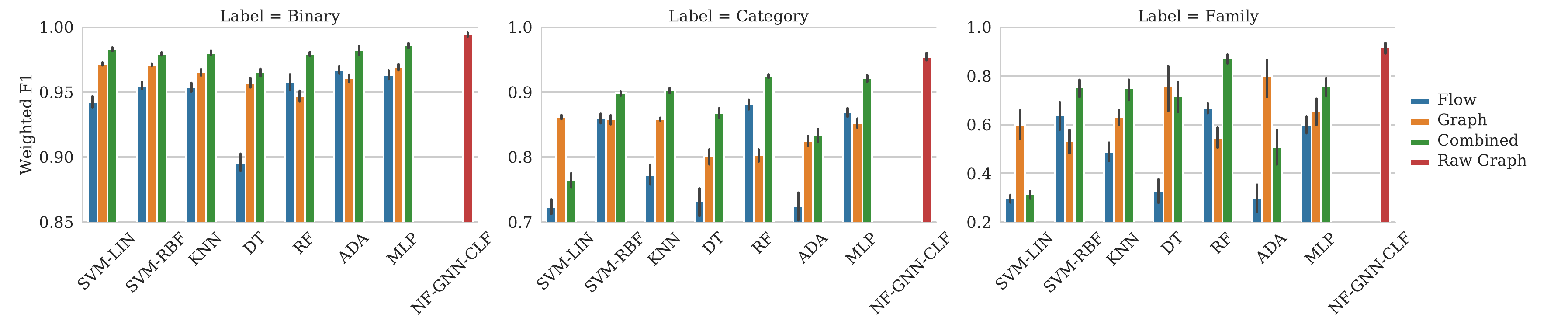}
	\caption{Weighted F1 scores for the three supervised tasks using different feature sets. Results are reported in terms of mean and standard deviation over 30 independent data splits.}
	\label{fig:supervised_feature_types}
\end{figure*}

\begin{figure}[t]
	\centering
	\includegraphics[width=\columnwidth]{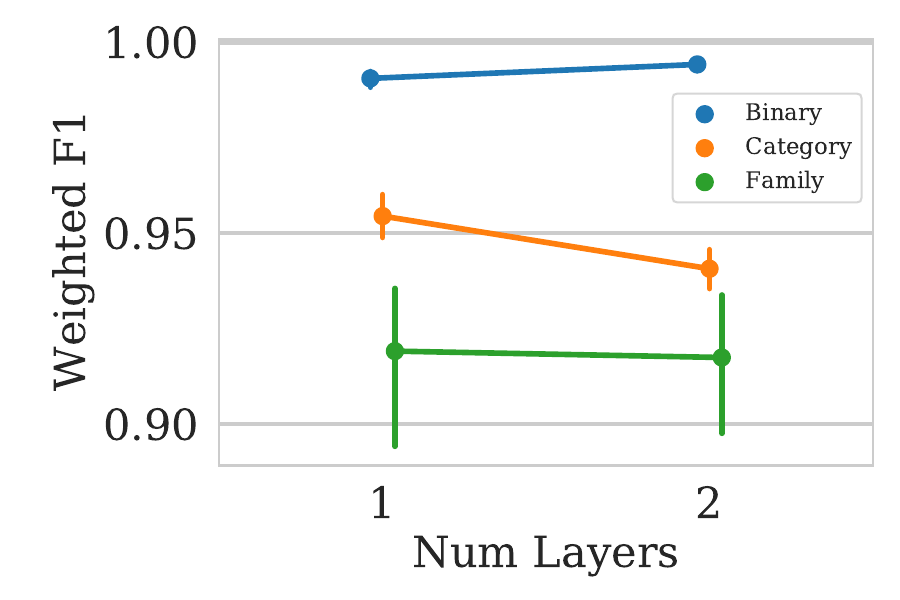}
	\caption{Performance of our model on supervised tasks with different numbers of layers.}
	\label{fig:supervised_num_layers}
\end{figure}

\subsubsection{Detection and Classification Performance}

Detection and classification results for all three tasks are reported in terms of weighted recall and precision in Table \ref{tab:results_supervised}.
For each of our baseline model, the best feature set (Flow, Graph, or Combined) has been chosen based on validation set performance. 

While some baselines perform notably weaker than others, we can observe that all baseline models still exceed the best results reported for flow classification in \citep{lashkari2018toward} in terms of both recall and precision, in most cases by a large margin.

Flow classification with an added static detection model and additional API-call features \cite{taheri2019extensible} performs better than some of our weaker baselines, especially in terms of recall. However, this approach is outperformed in terms of precision by all our baselines on category and family prediction. Our stronger baselines can outperform this approach, sometimes even by a large margin, except in terms of recall for family classification. It is important to note that \cite{taheri2019extensible} use additional data compared to our baselines and our model. Our approach could, in principle, be extended to also use this additional data and thus further boost performance.

Conversation-level detection \cite{abuthawabeh2020enhanced} performs worse than \cite{taheri2019extensible}, except for family classification. This competitor outperforms all of our baselines in terms of recall on the family classification task. In terms of precision, however, all of our baselines outperform all competitors by a significant margin.

The competitive and sometimes even vastly superior performance of our baselines already supports the main motivation of our approach to detect and classify malware using network flow graphs. 
Notably, this performance has been achieved even under a rigorous evaluation protocol and using relatively small training sets. In comparison, the competitors have used training set sizes between 60\% and 80\%.

Our proposed graph neural network model NF-GNN-CLF can further boost performance significantly compared to the baselines.
It is able to significantly exceed the performance of all competitors on all three tasks in terms of both recall and precision. In particular, compared to the best competitor, recall can be improved by 4.12\%, 14.41\%, and 24.78\% for binary, category, and family classification, respectively. Precision is improved by 4.14\%, 12.84\%, and 26.41\%. We wish to note that in most cases, the best competitor \cite{taheri2019extensible} uses additional data that is not available to our baselines and model. Compared to the other two competitors, the performance gain is even more significant. The largest improvement is achieved for the family classification task with a performance increase of over $65\%$ compared to flow classification \cite{lashkari2018toward}.

\subsubsection{Importance of Different Feature Sets}

To get more insight into the importance of different feature sets for our baseline models, we compare their performance in terms of weighted F1 score on different feature sets in Figure \ref{fig:supervised_feature_types}. We can observe that almost all baseline models perform best on the combined flow and graph feature set.

As a notable exception, the linear SVM performs best on category and family classification using only the graph features, and adding flow features significantly hurts performance. However, even using only graph features, this baseline still performs worst among all baselines. Similarly, DT and ADA perform best using only graph features for family classification but are also outperformed by other baselines using the combined feature set.
In most cases, the combination boosts performance significantly over the best individual feature set. 

Our model operating on the raw graphs still outperforms all baselines.

\subsubsection{Influence of the Number of Network Layers}

To further examine the influence of the number of layers on our model's performance, we compare different choices for the supervised classification tasks in Figure \ref{fig:supervised_num_layers}. We can observe that modeling 2-hop interactions between endpoints can boost performance on the binary prediction tasks, while direct interactions are more crucial for the remaining two tasks. In general, performance remains relatively stable for different numbers of layers.

\begin{table*}[t]
	\centering
	\resizebox{\textwidth}{!}{%
		\begin{tabular}{llllllll|cc}
\toprule
{} & \textbf{OC-SVM-LIN} & \textbf{OC-SVM-RBF} &      \textbf{LOF} &      \textbf{KDE} &       \textbf{IF} &   \textbf{MLP-AE} &   \textbf{MLP-OC} & \textbf{NF-GNN-AE} & \textbf{NF-GNN-OC} \\
\midrule
\textbf{Flow    } &    $57.17 \pm 4.81$ &    $70.01 \pm 1.90$ &  $73.15 \pm 1.22$ &  $72.50 \pm 1.03$ &  $70.39 \pm 1.12$ &  $71.64 \pm 1.20$ &  $81.29 \pm 4.07$ &           $\lvert$ &           $\lvert$ \\
\textbf{Graph   } &   $54.60 \pm 19.32$ &    $67.19 \pm 2.89$ &  $58.57 \pm 5.12$ &  $91.79 \pm 0.66$ &  $90.73 \pm 1.31$ &  $89.56 \pm 0.77$ &  $94.00 \pm 1.32$ &   $95.34 \pm 0.85$ &   $\mathbf{96.75 \pm 1.22}$ \\
\textbf{Combined} &    $58.21 \pm 4.83$ &    $76.07 \pm 1.78$ &  $77.94 \pm 1.56$ &  $86.60 \pm 1.07$ &  $85.73 \pm 2.21$ &  $83.04 \pm 0.78$ &  $94.03 \pm 4.47$ &           $\lvert$ &           $\lvert$ \\
\bottomrule
\end{tabular}
	}
	\vspace{1em}
	\caption{AUROC scores for unsupervised malware detection in terms of mean and standard deviation over 30 independent data splits. Our models use the raw graphs as input instead of the features extracted for the baseline models.}
	\label{tab:results_unsupervised}
\end{table*}
\begin{figure*}[t]
	\centering
	\includegraphics[width=\textwidth]{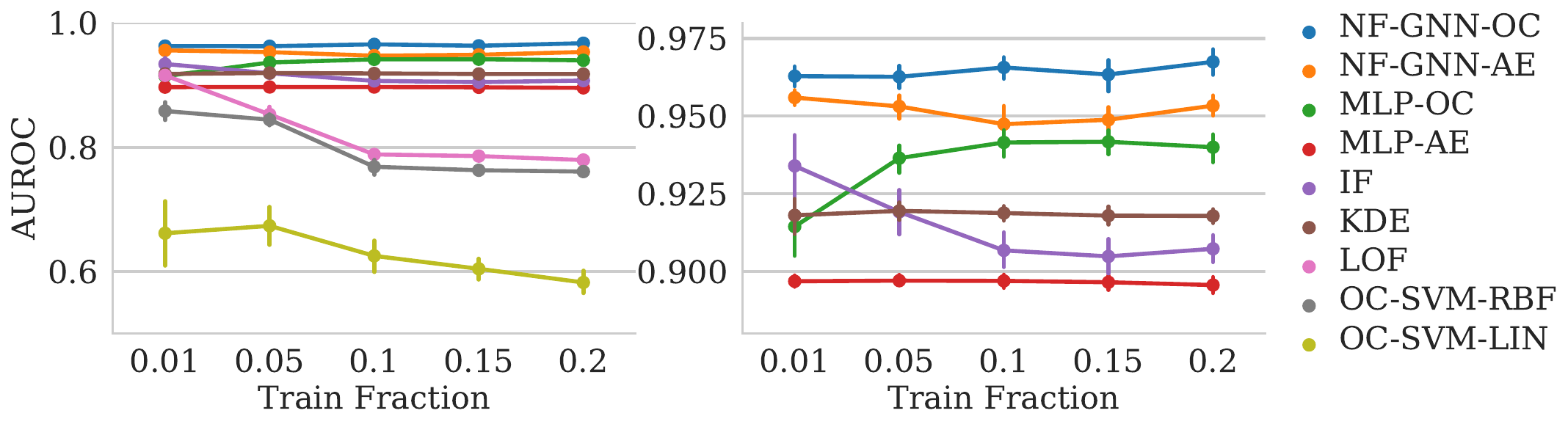}
	\caption{AUROC scores for unsupervised malware detection using different fractions of training samples. The right-hand figure provides an enlarged view of the upper part of the left-hand figure.}
	\label{fig:unsupervised_train_fraction}
\end{figure*}
\begin{figure}[t]
    \includegraphics[width=\columnwidth]{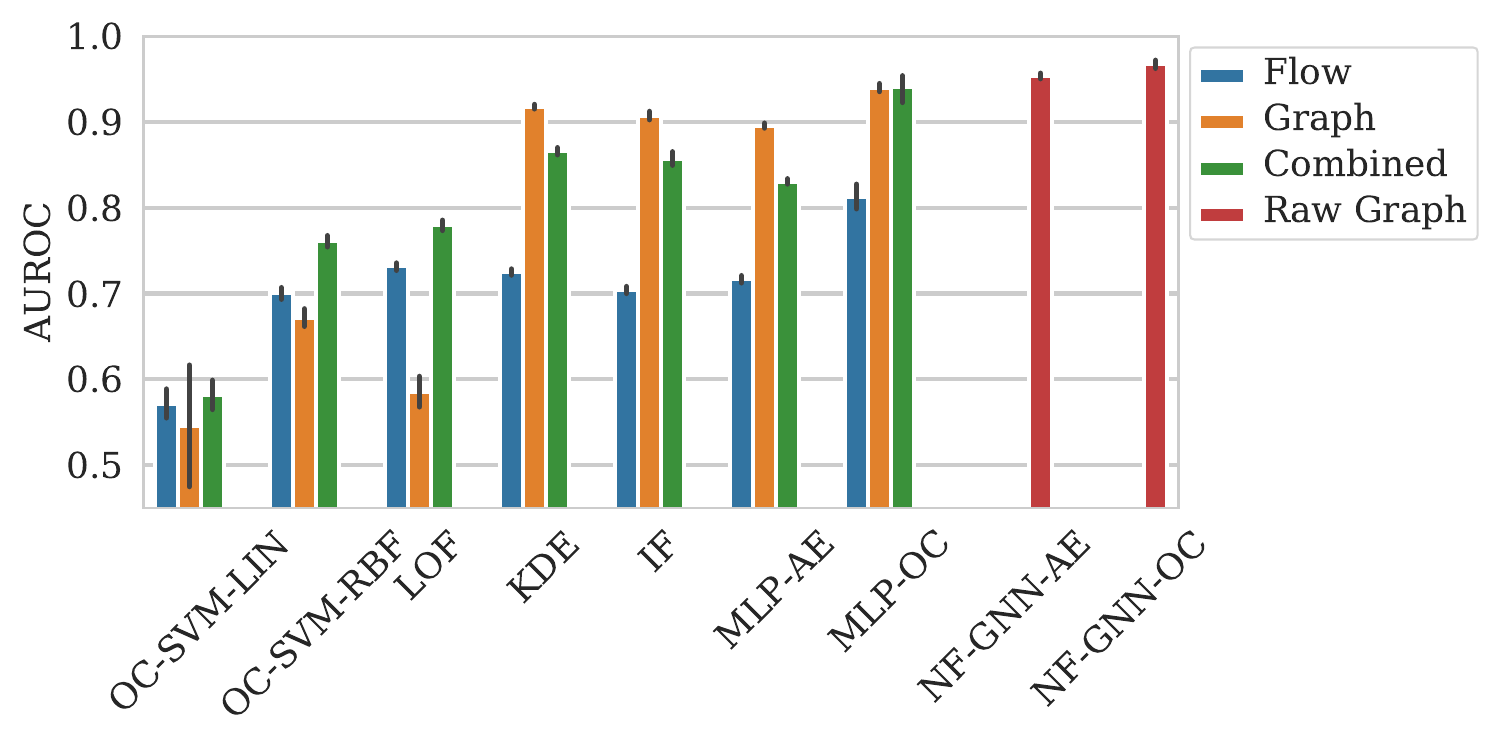}
    \caption{AUROC scores for unsupervised malware detection using different feature sets.}
    \label{fig:unsupervised_feature_types}
\end{figure}

\subsection{Unsupervised Malware Detection}

We further evaluate our approach in a more realistic unsupervised setting, where no labels are available for training. 

\subsubsection{Experimental Setup}

Our experimental setup is the same as in the supervised case with a few adjustments. To ensure a realistic distribution of benign and malicious samples, we perform stratified sampling to first split 20\% of the samples for training and then 10\% of the remaining samples for validation. The remaining samples are used for testing. All algorithms obtain access to the labeled validation set for hyper-parameter optimization, but training is still unsupervised. We evaluate detection performance using AUROC since it inherently adjusts for class imbalance \citep{campos2016evaluation}.

We evaluate against a set of popular baseline algorithms for anomaly detection, \emph{One-class SVM (OC-SVM)} with linear and RBF-kernel, \emph{Local Outlier Factor (LOF)}, \emph{Kernel Density Estimation (KDE)}, \emph{Isolation Forest (IF)}, \emph{Autoencoder with dense layers (MLP-AE)} and \emph{One-Class Neural Network with dense layers (MLP-OC)}. Again, all considered hyper-parameter values are provided in the supplement.
Since the flow- and conversation-based competitors \cite{lashkari2018toward,taheri2019extensible,abuthawabeh2020enhanced} only consider supervised detection and classification, we are not able to include them for comparison.

\subsubsection{Detection Performance}

Table \ref{tab:results_unsupervised} shows the detection results for different feature sets.
Results for different feature sets are additionally visualized in Figure \ref{fig:unsupervised_feature_types}. 
We can observe that several models report better prediction performance using only structural graph features. Thus, compared to the supervised setting, it is even more important to consider the topology of the network flow graph. While both neural network baselines, as well as KDE and IF already exhibit high detection performance, our models can again boost performance by a significant margin, demonstrating the importance of learning suitable representations from network flow graphs. We wish to emphasize that the performance of our unsupervised models, NF-GNN-AE and NF-GNN-OC, almost matches that of their supervised binary classification counterpart, NF-GNN-CLF, even without any labels provided for training.

\subsubsection{Influence of the Training Set Size}

In practice, there is often a shortage of available training data, even unlabeled data. Thus, we further investigate performance using different amounts of training data where we gradually reduce the fraction of training samples from 20\% to 1\%. For each training set size, a new grid search is performed to determine the best hyper-parameters for each model. Figure \ref{fig:unsupervised_train_fraction} shows that while some models perform worse with more available training data, possibly due to overfitting to anomalies in the training set, especially MLP-OC and NF-GNN-OC benefit from more training data. For all training set sizes, both of our models consistently outperform all competing baselines.

\section{Conclusion}
We proposed a novel network flow graph-based approach to malware detection and classification, where we monitor network traffic generated by a candidate application and extract a flow graph, which models communication between devices during the considered time frame. In addition, we proposed a novel edge feature-based graph neural network model along with three different model variants for supervised and unsupervised settings. Empirically, we found that even baseline models operating on manually extracted graph and flow features perform very well in all settings.
In addition, our proposed models automatically learn suitable representations of network flow graphs and can boost performance even further, significantly outperforming all competitors on all tasks.
Ablation studies examined the influence of different feature sets, the number of network layers, and training set size on detection and classification performance. In future work, we plan to consider additional network architectures, such as attention, model temporal dynamics, and consider explainability of model decisions.

\begin{acks}
This work was done during an internship at CT RDA BAM IBI-US, Siemens Technology, Princeton, NJ, USA.
\end{acks}

\balance
\bibliographystyle{ACM-Reference-Format}
\bibliography{bibliography}

\appendix
\section{Hyper-Parameter Optimization}
\label{app:hyper_parameters}

For the sake of a fair comparison, hyper-parameters of all models are optimized by a grid search based on performance on a separate validation set. For each algorithm, we consider a set of possible values for each of its adjustable hyper-parameters. While some common choices can be found in the literature, for the remaining hyper-parameters, we determine potential values, which seem most promising within the available computational budget. The considered hyper-parameter values for all supervised models can be found in Table \ref{tab:parameters_supervised}. The values considered for the unsupervised models are provided in Table \ref{tab:parameters_unsupervised}.

\begin{table*}[t]
    \centering
    \begin{tabular}{lll}
        \toprule
        \textbf{Algorithm} & \textbf{Parameter} & \textbf{Values} \\
        \midrule
         Support Vector Machine (SVM) & $C$ & $[2^{-7}, 2^{-6}, \dots, 2^{7}]$ \\
         & $\gamma$ (RBF-kernel) & $[2^{-7}, 2^{-6}, \dots, 2^{7}]$ \\
         k-Nearest Neighbor Classifier (KNN) & num. neighbors & $[1, 2, 3, 5, 8, 13, 21]$ \\
         Decision Tree (DT) & max. depth & $[2, 5, 10, None]$ \\
        & max. features & $[sqrt, None]$ \\
         Random Forest (RF) & num. estimators & $[10, 100, 1000]$ \\
        & criterion & $[entropy, gini]$ \\
        & max. features & $[sqrt, None]$ \\
         Adaboost (ADA) & num. estimators & $[10, 100, 1000]$ \\
        & learning rate & $[1e-3, 1e-2, 1e-1, 1]$ \\
        Multi-Layer Perceptron (MLP) & num. layers & $[1, 2]$ \\
        & num. hidden & $[16, 32, 64, 128]$ \\
        & L2-reg. & $[0, 1e-1, 1e-2, 1e-3, 1e-4]$ \\
        NF-GNN-CLF (ours) & num. layers & $[1, 2]$ \\
        & num. hidden & $[16, 32, 64, 128]$ \\
        & learning rate & $[1e-3, 1e-2]$ \\
        & dropout prob. & $[0, 0.2, 0.4, 0.6]$ \\
        & pool & $[mean, add, max]$ \\
        \bottomrule
    \end{tabular}
    \vspace{1em}
    \caption{Hyper-parameter values used in grid search for supervised algorithms.}
    \label{tab:parameters_supervised}
\end{table*}

\begin{table*}[t]
    \centering
    \begin{tabular}{lll}
        \toprule
        \textbf{Algorithm} & \textbf{Parameter} & \textbf{Values} \\
        \midrule
        One-class SVM (OC-SVM) & $\nu$ & $[1e-2, 1e-1]$ \\
        & $\gamma$ (RBF-kernel) & $[2^{-10}, 2^{-9}, \dots, 2^{10}]$ \\
        Local Outlier Factor (LOF) & num. neighbors & $[1, 2, 3, 5, 8, 13, 21]$ \\
        Kernel Density Estimation (KDE) & bandwidth & $[2^{0.5}, 2, \dots, 2^{5}]$ \\
        Isolation Forest (IF) & num. estimators & $[10, 100, 1000]$ \\
        & max. features & $[256, None]$ \\
        Autoencoder (MLP-AE) & num. layers & $[1, 2]$ \\
        & num. hidden & $[16, 32, 64, 128]$ \\
        & L2-reg. & $[0, 1e-1, 1e-2, 1e-3, 1e-4]$ \\
        One-class MLP (MLP-OC) & num. layers & $[1, 2]$ \\
        & num. hidden & $[16, 32, 64, 128]$ \\
        & L2-reg. & $[0, 1e-1, 1e-2, 1e-3, 1e-4]$ \\
        NF-GNN-AE (ours) & num. layers & $[1, 2]$ \\
        & num. hidden & $[16, 32, 64, 128]$ \\
        & learning rate & $[1e-3, 1e-2]$ \\
        & dropout prob. & $[0, 0.2, 0.4, 0.6]$ \\
        & pool & $[mean, add, max]$ \\
        NF-GNN-OC (ours) & num. layers & $[1, 2]$ \\
        & num. hidden & $[16, 32, 64, 128]$ \\
        & learning rate & $[1e-3, 1e-2]$ \\
        & dropout prob. & $[0, 0.2, 0.4, 0.6]$ \\
        & pool & $[mean, add, max]$ \\
        \bottomrule
    \end{tabular}
    \vspace{1em}
    \caption{Hyper-parameter values used in grid search for unsupervised algorithms.}
    \label{tab:parameters_unsupervised}
\end{table*}

\end{document}